\documentclass[preprint,12pt]{elsarticle}
\usepackage{amssymb,amsmath,paralist}
\usepackage{xfrac}
\usepackage{lipsum}
\usepackage[english]{babel}
\usepackage[utf8]{inputenc}
\usepackage{graphicx}
\usepackage[colorinlistoftodos]{todonotes}
\usepackage{amsfonts}
\usepackage{mathtools}
\usepackage{etoolbox}
\usepackage{mdframed}
\usepackage{enumitem}
\usepackage{algorithm}
\usepackage{algorithmicx}
\usepackage[noend]{algpseudocode}
\usepackage{bbm}
\usepackage{multirow}
\usepackage{rotating}
\usepackage{float}

\begin{document}

\begin{frontmatter}



\title{Stochastic mean-shift clustering}


\author{Itshak Lapidot}
\ead{itshakl@afeka.ac.il$^{1,2}$}
\address{
  $^1$Afeka Tel-Aviv Academic College of Engineering, ACLP, Israel\\
  $^2$Avignon University, LIA, France
  }

\begin{abstract}
In this paper we presented a stochastic version mean-shift clustering algorithm.  In the stochastic version the data points ''climb'' to the modes of the distribution collectively, while in the deterministic mean-shift, each datum ''climbs'' individually, while all other data points remains in their original coordinates. Stochastic version of the mean-shift clustering is comparison with a standard (deterministic) mean-shift clustering  on synthesized 2- and 3-dimensional data distributed between several Gaussian component. The comparison performed in terms of cluster purity and class data purity. It was found the the stochastic mean-shift clustering outperformed in most of the cases the deterministic mean-shift.

\end{abstract}


\end{frontmatter}

\section{Introduction}
\label{sec:Introduction}
Mean-shift clustering algorithm is a non-parametric iterative clustering algorithm. It estimates the probability density function of a random variable \cite{Fukunaga1975}. The clustering algorithm is applied to a variety of areas, like segmentation images, \cite{Tao2007, Paris2007}, particularly medical and satellite images \cite{Lu2011, Ai2014, Wu2015, Banerjee2012}, videos \cite{Wang2004},  and also applied to high dimensional data clustering \cite{Chakraborty2021}. An adapted version of mean-shift clustering was applied to short segments speaker clustering  \cite{Salmun2016a,Salmun2016b,Salmun2017,Cohen2021}.  This algorithm is deterministic and in an iterative procedure estimates the multi-modal \textit{probability density function} (\textit{pdf}) via the ''climbing'' path of each datum to its mode in a multi-modal distribution. All the data points that reached the same mode are grouped to the same cluster. We named it ''deterministic'' as the algorithm is indifferent to the order of the data that are chosen to find their modes, as at the end of the ''climbing'', each datum is placed back to its original place. There is a selective version of the mean-shift that chooses only some of the data points in order to speed up the clustering \cite{Senoussaoui2014}. It is sub-optimal, however, although a sub-set of the data points is chosen randomly; after the end of the ''climbing'' each datum is placed at the original place. A version of collective ''climbing'' for Gaussian kernel known as Gaussian blurring mean-shift, \cite{Carreira-Perpin2006}. In the Gaussian blurring mean-shift, the data points do not placed back to their original places, but collectively ''climb'' each time. This version is also deterministic but each new move depends on the current position of all the data points and not on the original their places.

In the current work we present a stochastic variant, where all the data points ''climb'' together, and after each single shift of one datum, a new datum is chosen randomly. The data are placed in their new positions and are not placed back to their original places. We will compare its performance versus the well-known deterministic mean-shift clustering. This will be done on several multi-model 2- and 3-dimensional distributions of synthesized data.

In this work is assumed that the problem is well defined - each cluster should represent a different Gaussian, while at the same time, each Gaussian must be concentrated in a single cluster. As such, a subjective, external evaluation is applied, i.e, a criterion that uses the true  labels of the data. We applied two similar, but different criteria. The first is a based on the quality/purity of a cluster, \cite{Manning2008} taking the value of the class with the maximum intersection with a given cluster, i.e., the winner takes all. We add in the similar way the quality/purity of the class, i.e. the maximal intersection between the class with all clusters. The second criterion is more smooth an represent the contribution of all the classes to each cluster and vice versa. This criterion is commonly used to validate the speaker clustering and diarization quality  \cite{Ajmera2002, Salmun2016a, Salmun2016b, Salmun2017, Valente2004, Anguera2006b, Dabbabi2017} and will be discussed in subsection \ref{subsec:ClusteringEvaluationCriteria}. First the \textit{average cluster purity} ($ACP$), and the \textit{average speaker purity} ($ASP$) are calculated, and then the $K$ value is evaluated as the geometric mean between the first two measures. In \cite{Lapidot2023} this criterion used for speaker clustering while compared the adapted versions of deterministic and stochastic mean-shift algorithms. The stochastic mean-shift always outperformed the deterministic mean-shift clustering.

The paper is organized as follows. In Section \ref{sec:MeanShiftAlgorithms}, clustering algorithms are presented, both deterministic and stochastic mean-shifts. Clustering evaluation criteria are also explained. Synthesized data are described is described in Section \ref{sec:DatasetsDesign}. At Section \ref{sec:ExperimentsResults}, the experiments with synthesized data are presented. The conclusions and the discussion are in Section \ref{sec:DiscussionConclusions}.

\section{Mean-shift algorithms}
\label{sec:MeanShiftAlgorithms}
\begin{figure}[!t]
    \centerline{\includegraphics[width=\columnwidth]{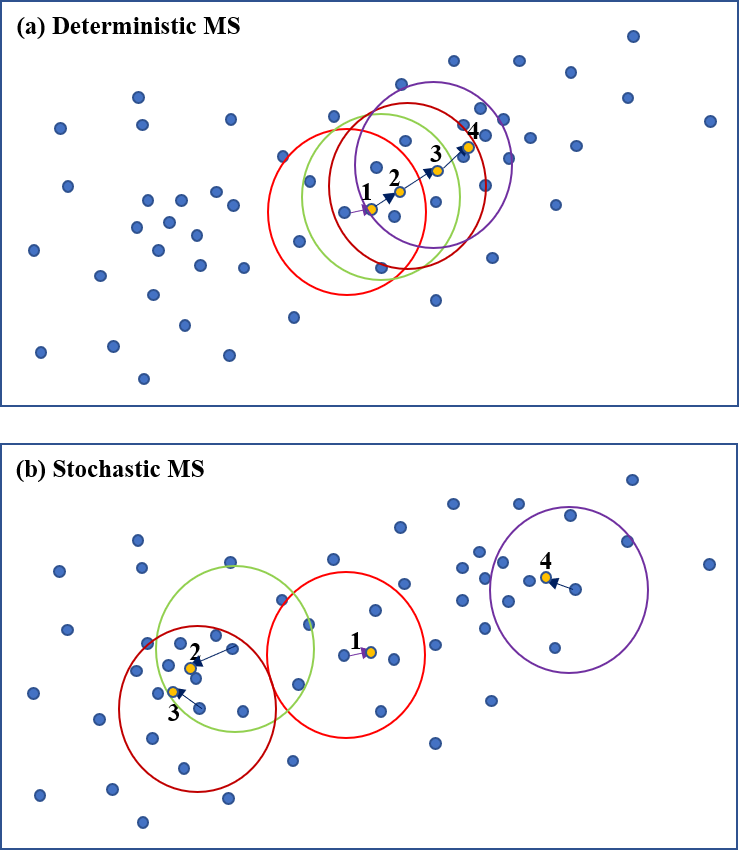}}
    \caption{\label{fig:MeanShift} Comparison between (a) deterministic mean-shift and (b) stochastic mean-shift.} 
    
\end{figure}

This section describes and evaluates the two mean-shift algorithms applied in this study. The difference between the deterministic mean-shift and the stochastic mean-shift is illustrated in Figure \ref{fig:MeanShift}. All the algorithms are variants of the mean-shift algorithm that is based on Euclidean distance \cite{Comaniciu2002} (subsection \ref{subsec:MeanShiftAlgorithms}, Figure \ref{fig:MeanShift}a). In this case, each time only one datum is shifted until convergence. Each time the neighbors of the same datum are found, and the shift of the datum is performed to the mean of the neighbors. This procedure is repeated until convergence. The final place is recorded and the datum returned to the original place, and the same procedure repeated with the next datum. This is done until all the data points are shifted. The stochastic version is then described in subsection \ref{subsec:stochsticMS}, Figure \ref{fig:MeanShift}b. In this case, each time another datum is randomly chosen. The neighbors are found, and the datum is shifted. The data points are never placed back in the original place, but participate in the finding for the neighborhood according to the shifted places (current places). At each iteration another datum is shifted, until the collective convergence of all the dataset.

\subsection{Mean-shift algorithm}
\label{subsec:MeanShiftAlgorithms}
The mean-shift algorithm is a non-parametric iterative algorithm. It estimates the \textit{probability density function} (\textit{pdf}) of a random variable \cite{Fukunaga1975}. The algorithm is inspired by the Parzen window approach to non-parametric density estimation. The algorithm does not require any prior knowledge regarding the number of clusters, and  assumes that the shape of each cluster is unimodal. Dense regions in the data space correspond to local maxima or to the \textit{pdf} modes. As such, for each data point, in order to reach the local maximum of the \textit{pdf}, a gradient ascent on the estimated local density is performed until convergence is reached. Each stationary point represents a mode of the density function. Data points that are associated with the same stationary point are assigned to the same cluster.

The gradient of the density function is required in order to find the above-mentioned modes. Following the mathematical formulation in \cite{Senoussaoui2014,Comaniciu2002,Fukunaga1975}, the mean shift vector ${m_h}(x )$ expression is derived according to Eq. (\ref{MeanShiftVectorFinal}).

\begin{equation}
\label{MeanShiftVectorFinal}
{m_h}(x ) = \frac{{\sum\limits_{{x _i} \in {S_h}\left( x  \right)} {{x _i}} }}{\# S_h\left(x\right)} - x
\end{equation}

where $x$ is the current position of the $d$-dimensional datum; $h$ is the neighborhood (or the bandwidth) from which the gradient is estimated, $\#$ donates the cardinality of the set of neighbor points, and ${S_h}\left( x  \right)$ is the set of data points that are the neighbors of datum $x$, ${S_h}(x ) \equiv \left\{ {{x_i}:\left\| {x  - {x_i}} \right\| \le h} \right\}$.

Let ${{\cal X}} = \left\{ {{x_j}} \right\}_{j = 1}^J$ be $d$-dimensional data to be clustered, then the mean-shift algorithm is as described in Algorithm \ref{alg:MeanShiftClustering}. For each $x_j$ from the set ${{\cal X}}$, an iterative procedure is applied: find the neighbors of $x_j$; calculate the shift according to Eq. (\ref{MeanShiftVectorFinal}); shift $x_j$; repeat the procedure with the shifted datum until convergence. Convergence in this case is defined as a shift's norm which is smaller than a threshold $Th_1$. The last step is merging the shifted data points that have a Euclidean distance smaller than $Th_2$ to the same cluster.

\begin{algorithm}
\caption{Mean-shift clustering algorithm}
\label{alg:MeanShiftClustering}
	\begin{algorithmic}
		\Require 
        \Statex A set of vectors, ${{\cal X}} = \left\{ {{x_j}} \right\}_{j = 1}^J$
 		\Comment{$x \in {\mathbb{R}^{d \times 1}}$}
 		\Statex Neighborhood size $h$
        \Comment{$h \in {\mathbb{R}^+}$}
        \Statex A cluster shift threshold $Th_1$
        \Comment{$Th_1 \in {\mathbb{R}^+}$}
        \Statex A cluster merging threshold $Th_2$
        \Comment{$Th_2 \in {\mathbb{R}^+}$}
		\For{$j:=1$ \textbf{to} $J$ \textbf{step} $1$}
            \State Set ${\hat x_j} = {x _j}$.
        	\Repeat
                \State Find the subset $S_h\left(\hat x _j\right)$.
               	\State Calculate $m_h\left(\hat x_j \right)$, the shift of the vector $\hat x_j$, using Eq. (\ref{MeanShiftVectorFinal}).
                \State ${\hat x _j} \leftarrow {\hat x _j} + {m_h}\left( {{\hat x _j}} \right)$
            \Until{$\left|m_h\left(x\right)\right|>Th_1$}
		\EndFor
		\State Cluster the shifted vectors $\hat {\cal X}=\left \lbrace \hat x_j \right \rbrace_{j=1}^J$ such that the distance between $2$ shifted vectors will be less then $Th_2$. 
	\Statex \textbf{Return:} Cluster index of each vector.
\end{algorithmic}
\end{algorithm}

\subsection{Stochastic mean-shift algorithm}
\label{subsec:stochsticMS}
In this section, a stochastic version of the mean-shift is presented. Two main differences exist between the deterministic and the stochastic algorithms:
\begin{enumerate}
    \item At each iteration, a new datum is taken randomly (with uniform probability), with repetitions, from the pool, while in the deterministic case, the same datum is iterated until convergence and then a new datum is taken. If the number of iterations is significantly larger than the size of the dataset, then with high probability, most (if not all) of the data points will be chosen several times and as such will probably converge to local optimum. Data points which will rarely be chosen will probably end as outlier clusters. In the experimental sections of this paper, it will be shown that the stochastic mean-shift frequently produces more clusters than the deterministic mean-shift. This might be the reason for this phenomenon.
    \item The neighbors that are chosen for the shift are taken from the data points in their current positions and not the original positions, as in the deterministic mean-shift. In this way, all the data paints ''climb'' collectively to their modes. With time, when the data points are shifted, the temporal density of the points should be greater at the modes, and as a result, should shift each next datum chosen in the more accurate direction.
\end{enumerate}
Algorithm \ref{alg:stochasticMeanShiftClustering} is the stochastic version of the mean-shift of Algorithm \ref{alg:MeanShiftClustering}.

The term ''convergence'' is present in both deterministic and stochastic algorithms. While ''convergence'' in the deterministic case is straightforward, i.e., iterate the same datum until the shift is smaller than a pre-defined threshold (or pre-defined number of iterations), in the stochastic case it must be defined more carefully. The use in this paper of the term ''convergence'' is: either the shift is smaller than a pre-defined threshold for a pre-defined number of iterations (as at each iteration a different datum is shifted); or there is a global pre-defined number of iterations for the entire procedure. If the number of iterations is sufficiently large and the data points are randomly taken according to a uniform distribution, then it is probable that most of the data points are shifted a sufficient number of times and it reach a place sufficiently close to treated as converged to a local optimum.

\begin{algorithm} [hbt!]
\caption{Stochastic mean-shift clustering algorithm}
\label{alg:stochasticMeanShiftClustering}
	\begin{algorithmic}
		\Require 
        \Statex A set of vectors, ${{\cal X}} = \left\{ {{x_j}} \right\}_{j = 1}^J$
 		\Comment{$x \in {\mathbb{R}^{d \times 1}}$}
 		\Statex Neighborhood size $h$
        \Comment{$h \in {\mathbb{R}^+}$}
        \Statex A cluster shift threshold $Th_1$
        \Comment{$Th_1 \in {\mathbb{R}^+}$}
        \Statex A cluster merging threshold $Th_2$
        \Comment{$Th_2 \in {\mathbb{R}^+}$}
        \Repeat
            \State Randomly choose ${\hat x_j} = {x_j}$.
            \State Find the subset $S_h\left(\hat x _j\right)$.
            \State Calculate $m_h\left(\hat x_j \right)$, the shift of the vector $\hat x_j$, using Eq. (\ref{MeanShiftVectorFinal}).
            \State ${x _j} \leftarrow {\hat x _j} + {m_h}\left( {{\hat x _j}} \right)$
        \Until{Convergence}
		\State Cluster the shifted vectors ${\cal X}=\left \lbrace x_j \right \rbrace_{j=1}^J$ such that the distance between $2$ shifted vectors will be less then $Th_2$. 
	\Statex \textbf{Return:} Cluster index of each vector.
\end{algorithmic}
\end{algorithm}

\subsection{Clustering evaluation criteria}
\label{subsec:ClusteringEvaluationCriteria}
There are two approaches evaluating clustering performance. The first is objective, internal evaluation, measuring clusters' compactness versus how far the clusters are one from the other. Parameters such as single-link, complete-link and many others are part of this approach, \cite{Manning2008, Bolshakova2003, Tibshirani2001, Chen2002evaluation}. These criteria do not assume any true labeling of the data, and as such it is task-independent. The second approach is task-dependent, external evaluation, and the clustering is compared with the ground-truth. In this work, external evaluation is applied.

One common way to evaluate clustering performances versus the ground-truth is using the purity of the obtained clusters. Evaluation criteria are formulated according to Eq. (\ref{EvalCriterionIntersection}). The notation is as follows:

\begin{itemize}
\item $R$ - Number of data attributions, different classes,
\item $Q$ - Number of clusters,
\item $n_{qr}$ - Total number of data points in cluster $q$ which are associated with class $r$,
\item $n_{.r}$ - Total number of data points which are associated with class $r$,
\item $n_{q.}$ - Total number of data points in cluster $q$,
\item $N$ - Total number of data points to be clustered,
\item $\mathcal{D} = \left\{d_r \right\}_{r=1}^R$ - The partitioning of the ground-truth data attribution,
\item $\mathcal{C} = \left\{c_q \right\}_{q=1}^Q$ - The partitioning of the data according to the clustering.
\end{itemize}

\begin{equation}\label{EvalCriterionIntersection}
\begin{array}{*{20}{c}}
{Pur_{\mathcal{C}} = \frac{1}{N}\sum\limits_{q = 1}^Q \underset{d_r \in \mathcal{D}}{\max} \, {\left| c_q \cap d_r \right|} }\\
{Pur_{\mathcal{D}} = \frac{1}{N}\sum\limits_{r = 1}^R \underset{c_q \in \mathcal{C}}{\max} \, {\left| c_q \cap d_r \right|} }\\
{G = \sqrt {Pur_{\mathcal{C}} \cdot Pur_{\mathcal{D}}} }
\end{array}
\end{equation}

 Equation \ref{EvalCriterionIntersection} describes, in the upper expression, a well-known way to measure the quality or purity of clustering, \cite{Manning2008}. It measures, in each cluster, which class is most prevalent and sums it over all the clusters. At the end it normalizes it by the total number of data points to bound it in the range $\left[0,1\right]$. However, this can lead to a trivial solution, when the number of clusters is equal to the size of the data, $N$, and each cluster contains exactly one datum. For this reason also, we suggest also to measure the purity of the data classes in the second expression. The logic behind this is to prefer clustering which clusters all the data points of the same class to the same cluster. This also can lead to a trivial solution of one cluster only. The optimal is to combine the two measures and in the current work it is done as a geometrical mean in the last expression.
 
Another criterion is the one that is used in speaker diarization and speaker clustering applications. This criterion was presented in \cite{Ajmera2002}, and was used in many cases \cite{Valente2004, Anguera2006b, Dabbabi2017} for speaker clustering. We applied it for short segments speaker clustering \cite{Salmun2016a, Salmun2016b, Salmun2017} and for speaker clustering quality estimation \cite{Cohen2017, Cohen2021}. The idea, similar to the first criterion, is to measure the impact of the data distribution on each cluster and the impact of distribution of the class between the clusters. Unlike the previous criterion, where only the maximal intersection is taken into account, here all the intersections are taken into account. Evaluation criteria are formulated according to Eq. (\ref{EvalCriterion}), First, the \textit{average cluster purity} ($ACP$) and the \textbf{\textit{average class purity}} have to be calculated. In previous works, this was named as \textit{average speaker purity} ($ASP$) and we will follow this notation for consistency. $ACP$ measures how well a cluster is limited, on average, to only one speaker, while $ASP$ measures how a speaker is limited, on average, to only one cluster. The ideal case is when both $ACP$ and $ASP$ are equal to $1.0$. Like in the first case, the geometrical mean of $ACP$ and $ASP$ is applied as an evaluation criterion, $K$. 

\begin{equation}\label{EvalCriterion}
\begin{array}{*{20}{c}}
{\begin{array}{*{20}{c}}
{ACP = \frac{1}{Q}\sum\limits_{q = 1}^Q {\frac{1}{{n_{q.}^2}}}\sum\limits_{r = 1}^R {\left| c_q \cap d_r \right|}^2 = \frac{1}{Q}\sum\limits_{q = 1}^Q {{p_{q.}}} } &;&{{p_{q.}} = \sum\limits_{r = 1}^R {\frac{{n_{qr}^2}}{{n_{q.}^2}}} }\\
{ASP = \frac{1}{R}\sum\limits_{r = 1}^R {\frac{1}{{n_{.r}^2}}}\sum\limits_{r = 1}^Q {\left| c_q \cap d_r \right|}^2 = \frac{1}{R}\sum\limits_{r = 1}^R {{p_{.r}}} }&;&{{p_{.r}} = \sum\limits_{r = 1}^Q {\frac{{n_{qr}^2}}{{n_{.r}^2}}} }
\end{array}}\\
{K = \sqrt {ACP \cdot ASP} }
\end{array}
\end{equation}

It is important to notice that $ACP$ is based on  cluster purities $\left\{ {{p_{q.}}} \right\}_{q = 1}^Q$ while $ASP$ is based on the speaker purities $\left\{ {{p_{.r}}} \right\}_{q = 1}^Q$. In both cases, the purities are not probabilities and they do not sum to one. As the criteria $ACP$ and $ASP$ are averages over purities and the information about the class or cluster sizes is omitted, it is biased toward the small classes/clusters. We will maintain this formulation in order to be consistent with previous work.

The values of the cluster purities $\left\{ {{p_{q.}}} \right\}_{q = 1}^Q$ and the speaker purities $\left\{ {{p_{.r}}} \right\}_{q = 1}^Q$ are closely related to the multi-class Gini measure \cite{Breiman1984}, and actually can be seen as $1-Gini\left(p_{q.}\right)$ and $1-Gini\left(p_{.r}\right)$ respectively.

There are advantages and disadvantages to both criteria. The disadvantage of one is the advantage of the other and vice versa. The advantage of the criterion $G$, which is based on $Pur_\mathcal{C}$ and $Pur_\mathcal{D}$, is in the fact that it gives a portion of the ''\textit{correct}'' data points from all the clusters, i.e.,\ at the end it is all divided by the dataset size $N$. For ACP and ACP, their values depend on averaging the purities of all the clusters, i.e.,\ all clusters have the same influence, independent of their sizes. The disadvantage of the $G$ criterion is that it takes into account only the size of the largest class in the cluster without taking into account the distribution of other classes in the cluster. It is not the same, if in addition to the largest class, there is only one additional large class in the cluster or many small classes. Similarly whether one class splits between two clusters or many clusters makes a difference. The $K$ criterion takes this into account, as $ASP$ and $ACP$ are calculated according to the class and cluster distributions and it is not just the majority which is the only important quantity in $Pur_\mathcal{C}$ and $Pur_\mathcal{D}$.

\section{Synthesized Datasets}
\label{sec:DatasetsDesign}

In this section, the data designed for comparison between the standard (deterministic) mean-shift and the stochastic mean-shift algorithms are presented. 

As mean-shift clustering is an estimate of the multi-modal probability density by iterative ”climbing” of each datum to its mode, several multi-modal datasets were designed. Synthesized datasets include $2$- and $3$-dimensional multi-modal distributions. Datasets are  differ by their distributions in several parameters. Mixture of Gaussian multivariate classes are defer in:
\begin{itemize}
\item Data dimension - $d$,
\item Number of classes (Gaussian mixtures) - $R$,
\item Amount of data per class - $n_{.r}$,
\item The mean vector of each class $\eta_r$,
\item The covariance matrix of each class $\Sigma_r$.
\end{itemize}

Seven datasets were designed:

\begin{enumerate}[label=\textbf{Set \arabic*:}]
\item Three 2 dimensional classes with:

\begin{equation*}\label{eq:set1}
\begin{array}{*{20}{c}}
{\eta_1 = \left[ {\begin{array}{*{20}{c}}
  1 \\ 
  1 
\end{array}} \right]} &
{\eta_2 = \left[ {\begin{array}{*{20}{c}}
  -1 \\ 
  -1 
\end{array}} \right]} &
{\eta_3 = \left[ {\begin{array}{*{20}{c}}
  1 \\ 
  -1 
\end{array}} \right]} &
{\Sigma_r = \left[ {\begin{array}{*{20}{c}}
  0.36 & 0.00 \\ 
  0.00 & 0.36 
\end{array}} \right]}
\end{array}
\end{equation*}

same isotropic covariance matrix for all classes and $n_{.r}=250$ for all classes,

\item same as \textbf{Set 1}, but:

\begin{equation*}\label{eq:set2}
{\Sigma_r = \left[ {\begin{array}{*{20}{c}}
  0.64 & 0.00 \\ 
  0.00 & 0.64 
\end{array}} \right]}
\end{equation*}

Broader Gaussians, while the means are the same, should be harder to cluster.

\item Same as \textbf{Set 2}, but $n_{.r}=50$, i.e. $5$ times less data,
\item Same as \textbf{Set 2}, but $n_{.r}=1500$, i.e. $6$ times more data,
\item Same as \textbf{Set 2}, but $\left\{ n_{.r} \right\} = \begin{Bmatrix}100 & 300 & 20 \end{Bmatrix}$, i.e. \ unbalanced data,

\item The mean vectors are the same, however the covariance matrices are not all spherical, $n_{.r}=250$ for all classes:

\begin{equation*}\label{eq:set6}
\begin{array}{*{20}{c}}
{\Sigma_1 = \left[ {\begin{array}{*{20}{c}}
  0.64 & 0.00 \\ 
  0.00 & 0.64 
\end{array}} \right]} &
{\Sigma_2 = \left[ {\begin{array}{*{20}{c}}
  0.73 & 0.48 \\ 
  0.48 & 0.73 
\end{array}} \right]} &
{\Sigma_3 = \left[ {\begin{array}{*{20}{c}}
  1.09 & -0.60 \\ 
  -0.60 & 1.09 
\end{array}} \right]}
\end{array}
\end{equation*}

\item Four 3-dimensional classes with $\left\{ n_{.r} \right\} = \begin{Bmatrix}250 & 300 & 200 & 200 \end{Bmatrix}$:

\begin{equation*}\label{eq:set7mean}
\begin{array}{*{20}{c}}
{\eta_1 = \left[ {\begin{array}{*{20}{c}}
  1 \\ 
  1 \\
  1
\end{array}} \right]} &
{\eta_2 = \left[ {\begin{array}{*{20}{c}}
  -1 \\ 
  -1 \\
  0
\end{array}} \right]} &
{\eta_3 = \left[ {\begin{array}{*{20}{c}}
  1 \\ 
  -1 \\
  0
\end{array}} \right]} &
{\eta_4 = \left[ {\begin{array}{*{20}{c}}
  -2 \\ 
  2 \\
  2
\end{array}} \right]} 
\end{array}
\end{equation*}

\begin{equation*}\label{eq:set7cov}
\begin{array}{*{20}{c}}
{\Sigma_1 = \left[ {\begin{array}{*{20}{c}}
  0.64 & 0.00 & 0.00 \\ 
  0.00 & 0.64 & 0.00 \\
  0.00 & 0.00 & 0.64
\end{array}} \right]} &
{\Sigma_2 = \left[ {\begin{array}{*{20}{c}}
  0.74 & 0.50 & -0.20 \\ 
  0.50 & 0.77 & -0.31 \\
  -0.2 & -0.31 & 0.41
\end{array}} \right]} \\
{\Sigma_3 = \left[ {\begin{array}{*{20}{c}}
  1.09 & -0.60 & -0.24 \\ 
  -0.60 & 1.73 & 1.60 \\
  -0.24 & 1.60 & 1.64
\end{array}} \right]} &
{\Sigma_4 = \left[ {\begin{array}{*{20}{c}}
  0.77 & 0.58 & 0.73 \\ 
  0.58 & 0.56 & 0.46 \\
  0.73 & 0.46 & 0.78
\end{array}} \right]}
\end{array}
\end{equation*}
\end{enumerate}

\section{Experiments and results}
\label{sec:ExperimentsResults}

In this section we present the comparison between deterministic and stochastic mean-shift algorithms applied to a $2$- and $3$-dimensional synthesized data (subsection \ref{subsec:MultiModal}).

\subsection{Multi-modal data clustering}
\label{subsec:MultiModal}

\begin{figure}[!t]
    \centerline{\includegraphics[width=\columnwidth]{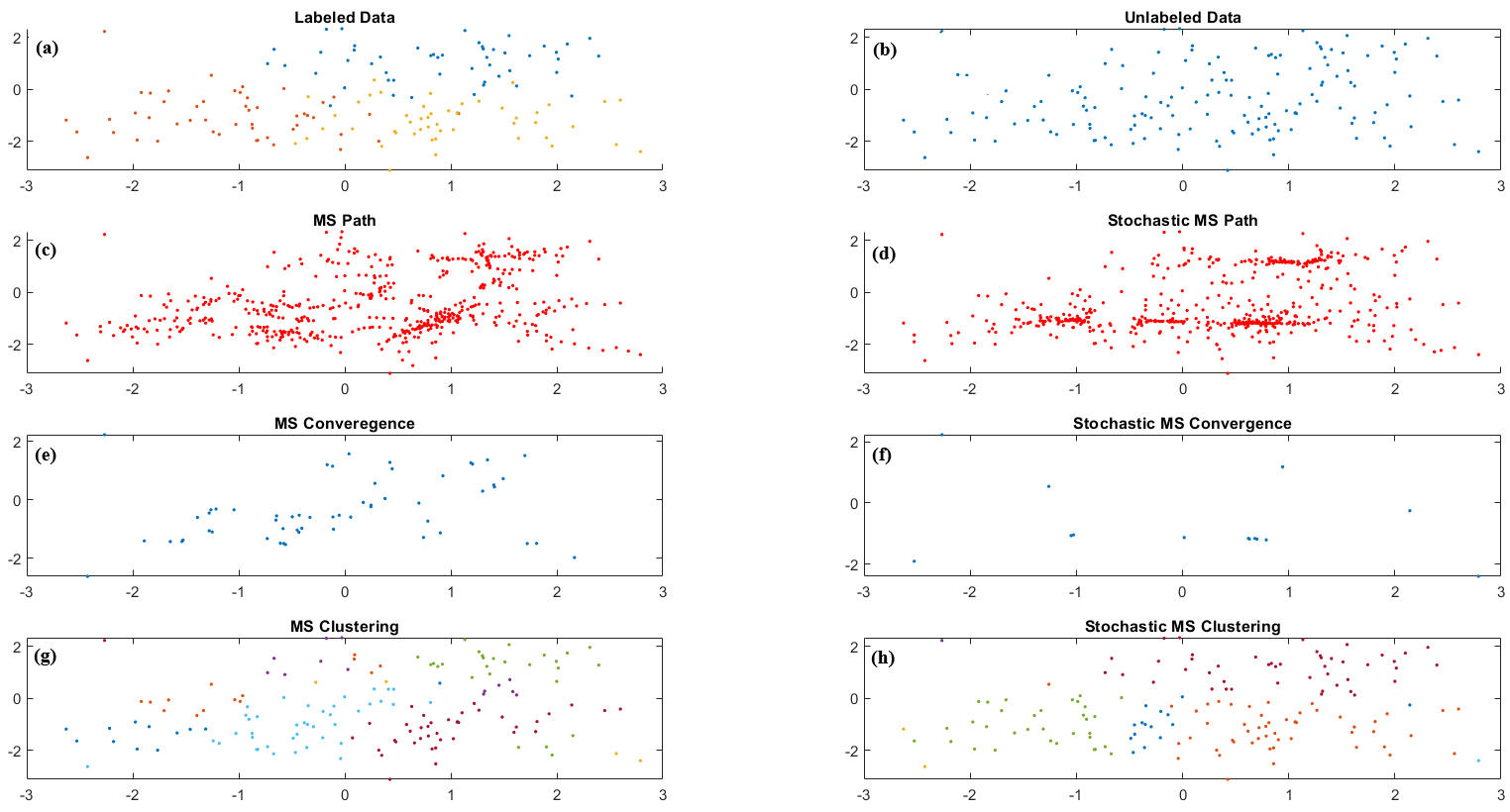}}
    \caption{\label{fig:ResultsExample} Results of Set 3 (a) data with true labels; (b) unlabeled data; (c) converging paths of the deterministic mean-shift; (d) converging paths of the stochastic mean-shift; (e) the found modes of the deterministic mean-shift; (f) the found modes of the stochastic mean-shift; (g) the clustering of the deterministic mean-shift; (f) the found clustering of the stochastic mean-shift.} 
\end{figure}

In this section we present the results on the datasets described in section \ref{sec:DatasetsDesign}. In Figure \ref{fig:ResultsExample} the clustering process is presented. Set 3 was taken as a test case. The true labels data and the unlabeled data are presented at sub-plots (a) and (b) respectively. Sub-plot (c) shows the paths of the data points during their shifts when the deterministic algorithm is applied, while sub-plot (d) shows the paths when the stochastic algorithm is applied. It can be seen that, in the deterministic case, the directed paths to the modes can be observed, while in the stochastic case they are much less visible. However, the high-density areas at the modes are clear. Sub-plots (e) and (f) shows the found modes of both algorithms. In this example there are far fewer  modes in the stochastic case. This does not mean that deterministic clustering results in more clusters, as at the last stage of the algorithm, close modes are merged into the same cluster. Sub-plots (g) and (h) show the final clusters of the deterministic and the stochastic algorithms respectively. In this example, deterministic mean-shift clustering ended with $14$ clusters while stochastic mean-shift clustering provided only $9$ clusters.

The results according to the datasets are as follows:

\begin{itemize}
\item \textbf{Set 1:} The dataset consists of the Gaussian mixtures with diagonal covariance matrix. All the variances are $0.36^2$ and $250$ data points per class. The results are summarized in Table\ref{table:Set1}. We can observe that using $K$ as evaluation criterion, the results of the stochastic algorithm is better than the deterministic, while according to $G$ as evaluation criterion they are almost the same. This can be explained as followed: $K$ takes into account the distribution of all classes in the cluster and not only the one that has the majority in the cluster to evaluate the cluster purity (\textit{ACP} versus $Pur_\mathcal{C}$), and the same for class purity (\textit{ASP} versus $Pur_\mathcal{D}$). In addition, the number of detected clusters is also provided.

\begin{table}[!ht]
   \footnotesize
   \centering
   \caption{\label{table:Set1}{The results of Set $1$ (\textbf{D} - deterministic mean-shift; \textbf{S} - stochastic mean-shift).}}
  \begin{tabular}{l||ccc|ccc|c|}
  	\cline{2-8}
    & \boldmath$ACP$ & \boldmath$ASP$ & \boldmath$K$ & \boldmath$Pur_\mathcal{C}$ & \boldmath$Pur_\mathcal{D}$ & \boldmath$G$ & \textbf{\#Clusters}\\
	\hline
	\multicolumn{1}{ |l|| }{\textbf{D}} & $0.78$ & $0.84$ & $0.81$ & $0.92$ & $0.92$ & $0.92$ & $4$\\
	\multicolumn{1}{ |l|| }{\textbf{S}} & $0.90$ & $0.82$ & $0.86$ & $0.92$ & $0.90$ & $0.91$ & $9$\\
	\hline
  \end{tabular}
\end{table}

\item \textbf{Set 2:} - It is similar to Set $1$, but the variances are $0.64^2$. As the means are the same in  Set $1$, the clustering problem becomes harder due to the larger overlap between the clusters. The results are summarized in Table \ref{table:Set2} and confirm the increase in clustering dificulty. Comparison of Tables \ref{table:Set1} and \ref{table:Set2} confirms the degradation, however in terms of $K$, the stochastic algorithm degrades more, for $G$ a little less.

\begin{table}[!ht]
   \footnotesize
   \centering
   \caption{\label{table:Set2}{The results of Set $2$ (\textbf{D} - deterministic mean-shift; \textbf{S} - stochastic mean-shift).}}
  \begin{tabular}{l||ccc|ccc|c|}
  	\cline{2-8}
    & \boldmath$ACP$ & \boldmath$ASP$ & \boldmath$K$ & \boldmath$Pur_\mathcal{C}$ & \boldmath$Pur_\mathcal{D}$ & \boldmath$G$ & \textbf{\#Clusters}\\
	\hline
	\multicolumn{1}{ |l|| }{\textbf{D}} & $0.85$ & $0.70$ & $0.77$ & $0.85$ & $0.82$ & $0.84$ & $8$\\
	\multicolumn{1}{ |l|| }{\textbf{S}} & $0.81$ & $0.72$ & $0.77$ & $0.87$ & $0.84$ & $0.86$ & $10$\\
	\hline
  \end{tabular}
\end{table}

\item \textbf{Set 3:} In this case we repeat the Set $2$ case but with five times less data per cluster (Figure \ref{fig:ResultsExample}).  The results are summarized in Table \ref{table:Set3}. In the deterministic case, the clustering ended with many clusters; each was pure, however, as the data spread across many clusters, the class purity is very low. In the stochastic case, there are much fewer clusters, approximately as for Set $2$ and the results are similar.

\begin{table}[!ht]
   \footnotesize
   \centering
   \caption{\label{table:Set3}{The results of Set 3 (\textbf{D} - deterministic mean-shift; \textbf{S} - stochastic mean-shift).}}
  \begin{tabular}{l||ccc|ccc|c|}
  	\cline{2-8}
    & \boldmath$ACP$ & \boldmath$ASP$ & \boldmath$K$ & \boldmath$Pur_\mathcal{C}$ & \boldmath$Pur_\mathcal{D}$ & \boldmath$G$ & \textbf{\#Clusters}\\
	\hline
	\multicolumn{1}{ |l|| }{\textbf{D}} & $0.92$ & $0.36$ & $0.57$ & $0.85$ & $0.53$ & $0.67$ & $14$\\
	\multicolumn{1}{ |l|| }{\textbf{S}} & $0.89$ & $0.65$ & $0.76$ & $0.88$ & $0.79$ & $0.84$ & $9$\\
	\hline
  \end{tabular}
\end{table}

\item \textbf{Set 4:} Here the data is still balanced, but with much more data points per class ($1500$). The results are summarized in Table \ref{table:Set4}. Two things are of great interest. First, the number of clusters in the deterministic mean-shift is much higher than in the stochastic mean-shift. We observed this phenomenon when we repeated this experiment several times. Second, there is a very large difference between the \textit{ACP} and the $Pur_C$ in the deterministic mean-shift. This can be explained when looking at the results in Figure \ref{fig:ResultsExample1500}. The data is very dense and in the deterministic case, as at the end of the convergence of one datum it is placed back to the original place before starting with the next datum, all the modes are very close to each other as can be seen in subplot \ref{fig:ResultsExample1500}e. This is not the situation in the stochastic case, as if the data points shifted and drifted apart, they are not placed back and each datum continues to move to a different mode (subplot \ref{fig:ResultsExample1500}f). This causes one large cluster during the merging and several outliers in the deterministic case (subplot \ref{fig:ResultsExample1500}g), versus \ref{fig:ResultsExample1500}h where we can see three well-separated clusters; one small cluster at the intersection of the other three, and all the others can be viewed as outliers. The \textit{ACP} calculation averages the cluster purity of each cluster, without taking into account the size of each cluster. This leads to a relatively high \textit{ACP} as four clusters of the outliers are pure and one is not. In contrast, $Pur_\mathcal{C}$ counts only the total number of well-clustered data points. This leads to a low value of $Pur_\mathcal{C}$. It is important to mention that the results of the stochastic mean-shift clustering almost unchanged as a function of the amount of the data per class, Sets $2 - 4$.

\begin{table}[!ht]
   \footnotesize
   \centering
   \caption{\label{table:Set4}{The results of Set $4$ (\textbf{D} - deterministic mean-shift; \textbf{S} - stochastic mean-shift).}}
  \begin{tabular}{l||ccc|ccc|c|}
  	\cline{2-8}
    & \boldmath$ACP$ & \boldmath$ASP$ & \boldmath$K$ & \boldmath$Pur_\mathcal{C}$ & \boldmath$Pur_\mathcal{D}$ & \boldmath$G$ & \textbf{\#Clusters}\\
	\hline
	\multicolumn{1}{ |l|| }{\textbf{D}} & $0.87$ & $1.00$ & $0.93$ & $0.33$ & $1.00$ & $0.58$ & $5$\\
	\multicolumn{1}{ |l|| }{\textbf{S}} & $0.88$ & $0.69$ & $0.78$ & $0.85$ & $0.82$ & $0.84$ & $20$\\
	\hline
  \end{tabular}
\end{table}

\begin{figure}[!t]
    \centerline{\includegraphics[width=\columnwidth]{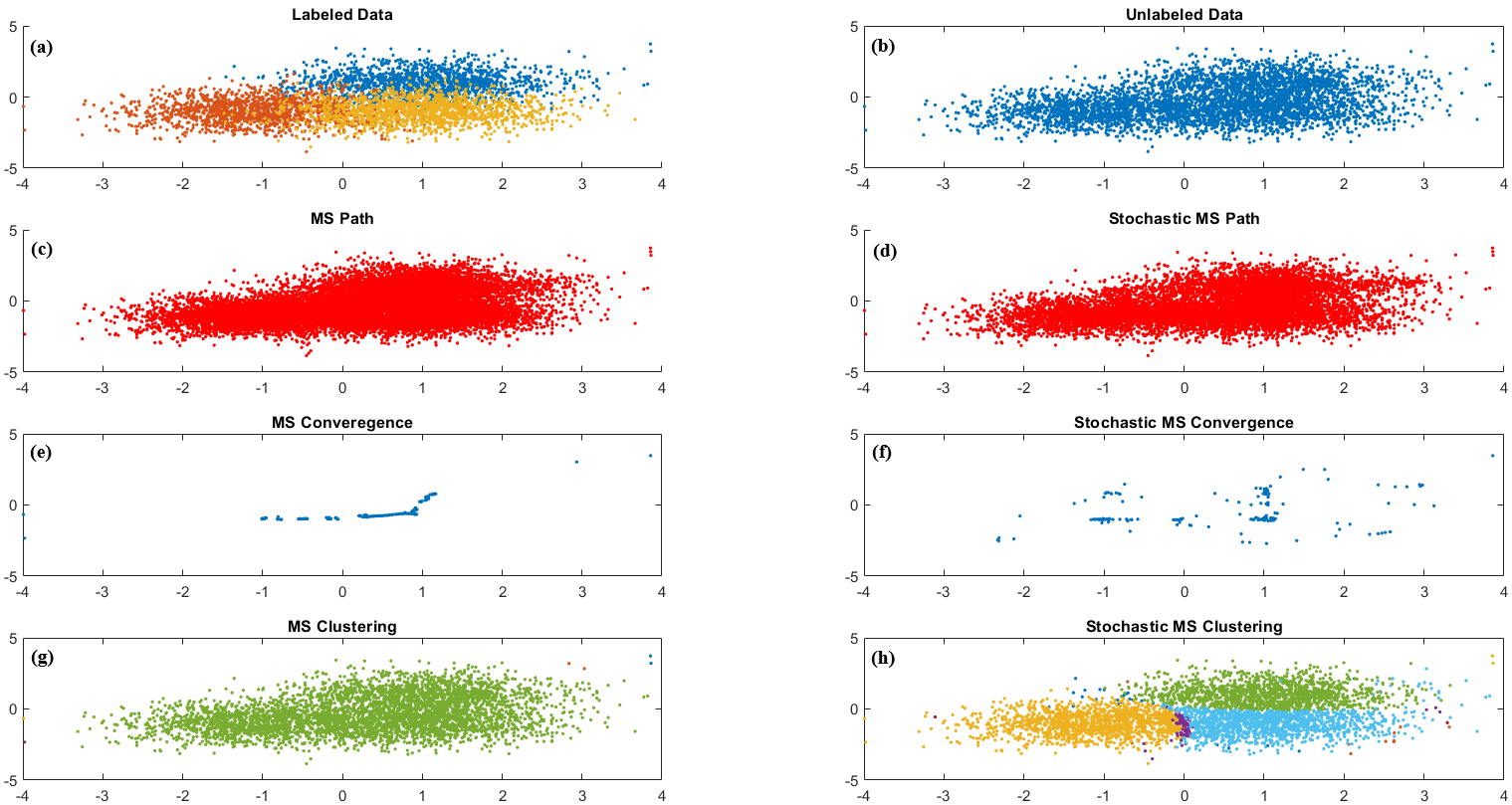}}
    \caption{\label{fig:ResultsExample1500} Results of Set $4$ (a) data with true labels; (b) unlabeled data; (c) converging paths of the deterministic mean-shift; (d) converging paths of the stochastic mean-shift; (e) the modes found of the deterministic mean-shift; (f) the modes found of the stochastic mean-shift; (g) the clustering of the deterministic mean-shift; (f) the found clustering of the stochastic mean-shift.} 
\end{figure}

\item \textbf{Set 5:} In this case the data was unbalanced, but the number of data points was not too large. The result are summarized in Table \ref{table:Set5}. From our observation, it is not conclusive which algorithm is better, similar to the results we obtained for Sets $1$ and $2$, Tables \ref{table:Set1} and \ref{table:Set2} respectively. The results become constantly better for the stochastic mean-shift when the amount of the data is increased by about an order: instead of $\left\{ n_{.r} \right\} = \begin{Bmatrix}100 & 300 & 20 \end{Bmatrix}$, cluster sizes were increased to $\left\{ n_{.r} \right\} = \begin{Bmatrix}1250 & 3750 & 250 \end{Bmatrix}$ ($12.5$ times more). This is due to the same effect as in the previous experiment with Set $4$.

\begin{table}[!ht]
   \footnotesize
   \centering
   \caption{\label{table:Set5}{The results of Set $5$ (\textbf{D} - deterministic mean-shift; \textbf{S} - stochastic mean-shift).}}
  \begin{tabular}{l||ccc|ccc|c|}
  	\cline{2-8}
    & \boldmath$ACP$ & \boldmath$ASP$ & \boldmath$K$ & \boldmath$Pur_\mathcal{C}$ & \boldmath$Pur_\mathcal{D}$ & \boldmath$G$ & \textbf{\#Clusters}\\
	\hline
	\multicolumn{1}{ |l|| }{\textbf{D}} & $0.95$ & $0.59$ & $0.75$ & $0.92$ & $0.82$ & $0.87$ & $9$\\
	\multicolumn{1}{ |l|| }{\textbf{S}} & $0.88$ & $0.69$ & $0.78$ & $0.94$ & $0.92$ & $0.93$ & $9$\\
	\hline
  \end{tabular}
\end{table}

\item \textbf{Set 6:} In this experiment the data was generated with non-diagonal covariance matrices with same cluster size, $250$. The results are summarized in Table \ref{table:Set6}. The stochastic algorithm ended with $2$ times more clusters and the much better fit to the same class (both according to \textit{ACP} and $Pur_\mathcal{C}$).  In total, the stochastic mean-shift performed more accurate.

\begin{table}[!ht]
   \footnotesize
   \centering
   \caption{\label{table:Set6}{The results of Set $6$ (\textbf{D} - deterministic mean-shift; \textbf{S} - stochastic mean-shift).}}
  \begin{tabular}{l||ccc|ccc|c|}
  	\cline{2-8}
    & \boldmath$ACP$ & \boldmath$ASP$ & \boldmath$K$ & \boldmath$Pur_\mathcal{C}$ & \boldmath$Pur_\mathcal{D}$ & \boldmath$G$ & \textbf{\#Clusters}\\
	\hline
	\multicolumn{1}{ |l|| }{\textbf{D}} & $0.71$ & $0.69$ & $0.70$ & $0.71$ & $0.78$ & $0.74$ & $5$\\
	\multicolumn{1}{ |l|| }{\textbf{S}} & $0.91$ & $0.65$ & $0.77$ & $0.77$ & $0.75$ & $0.76$ & $10$\\
	\hline
  \end{tabular}
\end{table}

\item \textbf{Set 7:} In the last experiment the data was $3$-dimensional with four clusters that have similar, but not the same, sizes (all between $200$ and $300$ data points). The results are summarized in Table \ref{table:Set7}. The performance of the stochastic mean-shift are slightly better than of the deterministic mean-shift. It seems that increasing both, the dimenssionality of the data and the number of clusters results that the number of ''\textit{outlier}'' clusters is increases. This can be tuned by increasing the threshold of merging the data point to clusters.

\begin{table}[!ht]
   \footnotesize
   \centering
   \caption{\label{table:Set7}{The results of Set $7$ (\textbf{D} - deterministic mean-shift; \textbf{S} - stochastic mean-shift).}}
  \begin{tabular}{l||ccc|ccc|c|}
  	\cline{2-8}
    & \boldmath$ACP$ & \boldmath$ASP$ & \boldmath$K$ & \boldmath$Pur_\mathcal{C}$ & \boldmath$Pur_\mathcal{D}$ & \boldmath$G$ & \textbf{\#Clusters}\\
	\hline
	\multicolumn{1}{ |l|| }{\textbf{D}} & $0.89$ & $0.47$ & $0.65$ & $0.87$ & $0.62$ & $0.73$ & $32$\\
	\multicolumn{1}{ |l|| }{\textbf{S}} & $0.92$ & $0.52$ & $0.69$ & $0.86$ & $0.68$ & $0.76$ & $24$\\
	\hline
  \end{tabular}
\end{table}
\end{itemize}

The summary of all the experiments is given in Table \ref{table:Summary}. The best results of each experiment according to each of the criteria ($K$ or $G$) are appears in \textbf{bold}. It can be seen that the stochastic mean-shift is mostly performed better than the deterministic mean-shift. Only for the Set $4$ for $K$ criterion the deterministic mean-shift gave a significantly higher score. It can be explained by the fact that the stochastic mean-shift produced $20$ clusters instead of actual $3$, while the deterministic mean-shift produced only $5$ clusters. In such a case, the \textit{ASP} for the in the stochastic case was relatively low. However, at Figure \ref{fig:ResultsExample1500} we see the reason of such a difference. In the deterministic case, almost all the data converged to a single cluster. It leads to a very high \textit{ASP}. The \textit{ACP} is not very low due to the fact that the outlier clusters are very pure.

\begin{table}[!ht]
   \footnotesize
   \centering
   \caption{\label{table:Summary}{The results of Set $7$ (\textbf{D} - deterministic mean-shift; \textbf{S} - stochastic mean-shift).}}
  \begin{tabular}{lcccccccccccccc}
    \cline{2-15}
    & \multicolumn{2}{|c||}{{\textbf{\textit{Set 1}}}}
    & \multicolumn{2}{c||}{{\textbf{\textit{Set 2}}}}
    & \multicolumn{2}{c||}{{\textbf{\textit{Set 3}}}}
    & \multicolumn{2}{c||}{{\textbf{\textit{Set 4}}}}
    & \multicolumn{2}{c||}{{\textbf{\textit{Set 5}}}}
    & \multicolumn{2}{c||}{{\textbf{\textit{Set 6}}}}
    & \multicolumn{2}{c|}{{\textbf{\textit{Set 7}}}}\\
    & \multicolumn{1}{|c}{\textbf{D}} & \multicolumn{1}{c||}{\textbf{S}} & \textbf{D} & \multicolumn{1}{c||}{\textbf{S}} & \textbf{D} & \multicolumn{1}{c||}{\textbf{S}} & \textbf{D} & \multicolumn{1}{c||}{\textbf{S}} & \textbf{D} & \multicolumn{1}{c||}{\textbf{S}} & \textbf{D} & \multicolumn{1}{c||}{\textbf{S}} & \textbf{D} & \multicolumn{1}{c|}{\textbf{S}}\\
	\hline \hline
 
    \multicolumn{1}{|c}{\boldmath$K$} & \multicolumn{1}{|c}{$0.81$} & \multicolumn{1}{c||}{\boldmath$0.86$} & \boldmath$0.77$ & \multicolumn{1}{c||}{\boldmath$0.77$} & $0.57$ & \multicolumn{1}{c||}{\boldmath$0.76$} & \boldmath$0.93$ & \multicolumn{1}{c||}{$0.78$} & $0.75$ & \multicolumn{1}{c||}{\boldmath$0.78$} & $0.70$ & \multicolumn{1}{c||}{\boldmath$0.77$} & $0.55$ & \multicolumn{1}{c|}{\boldmath$0.69$}\\
    
    \multicolumn{1}{|c}{\boldmath$G$} & \multicolumn{1}{|c}{\boldmath$0.92$} & \multicolumn{1}{c||}{$0.91$} & $0.84$ & \multicolumn{1}{c||}{\boldmath$0.86$} & $0.67$ & \multicolumn{1}{c||}{\boldmath$0.84$} & $0.58$ & \multicolumn{1}{c||}{\boldmath$0.84$} & $0.87$ & \multicolumn{1}{c||}{\boldmath$0.93$} & $0.74$ & \multicolumn{1}{c||}{\boldmath$0.76$} & $0.73$ & \multicolumn{1}{c|}{\boldmath$0.76$}\\
    \hline
  \end{tabular}
\end{table}

\section{Discussion and conclusions}
\label{sec:DiscussionConclusions}

In this work we presented a stochastic version of the mean-shift clustering algorithm. Originally, the mean-shift algorithm is deterministic. Each vector in the data ''climbs'' separately to one of the \textit{pdf} modes, while other vectors remain at their original places. Then, in turn, the same procedure is applied to another data point, and this procedure is repeated until all the vectors reach their modes. In the algorithm presented, for each iteration a new datum is chosen randomly, with replacement for a single shift only. The neighbors of each datum are found according to data points' current positions and not according to the original position (as in the deterministic case). In such a strategy, the data points converge to their modes jointly rather than one by one.

The original mean-shift clustering was compared with its stochastic version. Two- and three-dimensional data sets were synthesized with three and four Gaussian mixtures. Different scenarios were examined: balanced and unbalanced data, small ans large amounts of data per class and different covariance matrices. It can be seen that the stochastic algorithm performed as well, or better than,  the deterministic mean-shift (Tables \ref{table:Set1}-\ref{table:Set7}). The summary of the results is presented in Table \ref{table:Summary}. It seems that the stochastic mean-shift has an advantage when the amount of data is large and the modes becomes more blurred (Figure \ref{fig:ResultsExample1500} and Table \ref{table:Set4}, \textit{G} criterion). Frequently, the stochastic mean-shift produces more clusters than the deterministic mean-shift. Nevertheless, the performance is at least as good as in the deterministic case. This is because the extra clusters are small, and the result of data points that were rarely chosen (low density) and become outlier clusters.

Another observation relates to the number of estimated clusters. Both algorithms suffer from under-clustering and produce more clusters than the true number of classes. However, the stochastic algorithm produces more clusters than the deterministic mean-shift clustering in most of the experiments. This can be maintained by tuning the hyper-parameter of $Th_2$ which responsible for the merging procedure. Increase of $Th_2$ leads to smaller number of larger clusters., however, it does not lead to a better clustering.

In summary, in most of the examples, stochastic mean-shift clustering performed better or at least as good as the deterministic mean-shift clustering. In the future it will be tested on real data to compare these clustering algorithms.

%\section*{Acknowledgement}
% \linenumbers

%\section{}
%\label{}



%% References

\bibliographystyle{model5-names}\biboptions{authoryear}
\bibliography{mybib}
%\input appendix






% \end{thebibliography}

%\input FiguresAndTables

\end{document}